%% file: main.tex
\documentclass[11pt, letterpaper]{arxiv}

\usepackage[all]{hypcap}
\usepackage[comma,numbers,sort,compress]{natbib}
\usepackage[capitalize,noabbrev]{cleveref}
\usepackage{graphicx}
\usepackage{subfigure}
\usepackage{float}
\usepackage{wrapfig}
\usepackage{placeins}

\usepackage{booktabs}
\usepackage{colortbl}
\usepackage{threeparttable}
\usepackage{multirow}
\usepackage{tabularx}
\usepackage{fleqn, tabularx}

\usepackage{amsmath}
\usepackage{amssymb}
\usepackage{mathtools}
\usepackage{amsthm}

\usepackage{algorithm}
\usepackage{algpseudocode}

\usepackage{microtype}
\usepackage{setspace}
\usepackage{xcolor}
\usepackage[most]{tcolorbox}

\usepackage{enumitem}

\usepackage{listings}
\usepackage{color}

\definecolor{sb_blue}{RGB}{65,105,225}
\definecolor{sb_orange}{RGB}{255,140,0}
\definecolor{gray}{RGB}{128,128,128}

\lstdefinestyle{mystyle}{
    language=Python,
    xleftmargin=5.0ex,
    basicstyle=\footnotesize\ttfamily\linespread{4},
    backgroundcolor=\color{gray!10},
    commentstyle=\color{gray},
    alsoletter={<>-0123456789},
    keywordstyle=\color{sb_blue},
    ndkeywords={nn, F, torch, partial},
    ndkeywordstyle=\color{sb_orange},
    emph={import, def, return, if, else},
    emphstyle=\bfseries\color{sb_blue},
    numberstyle=\footnotesize\ttfamily\color{gray},
    stringstyle=\color{sb_blue},
    breakatwhitespace=false,
    breaklines=true,
    keepspaces=true,
    numbers=left,
    numbersep=5pt,
    showspaces=false,
    showstringspaces=false,
    showtabs=false,
    tabsize=2
}
\definecolor{lightblue}{rgb}{0.22,0.45,0.70}

\title{ExToken: Structured Exploration for Efficient Vision-Language-Action Reinforcement Fine-tuning}

\author[1,2]{Yilun Kong}
\author[2,3]{Yunpeng Qing}
\author[1]{Guozheng Ma}
\author[1]{Haoyu Wang}
\author[4]{Li Shen}
\author[2]{ \\ Zhi Hou}
\author[1,*]{Dacheng Tao}

\affil[1]{Nanyang Technological University}
\affil[2]{ACE Robotics}
\affil[3]{Zhejiang University}
\affil[4]{Sun Yat-sen University \par \vspace{-0.8em} {\footnotesize *}Corresponding author.}

\begin{abstract}
\input{Section/0_Abs}   
\end{abstract}

\begin{document}
\maketitle

\input{Section/1_Intro}
\input{Section/2_Rethink}    

\input{Section/3_Method}

\input{Section/4_Exp}

\input{Section/5_Conclusion}

\clearpage
\bibliography{ref}               

\newpage
\appendix
\onecolumn
\addtocontents{toc}{\protect\setcounter{tocdepth}{-1}}
\input{Section/Appendix}

\end{document}

%% file: Section/0_Abs.tex
Reinforcement Learning (RL) has demonstrated significant potential for improving Vision-Language-Action (VLA) models on complex manipulation tasks.  However, its practical scalability remains severely limited by the substantial cost of environmental interactions.
    In this work, we first investigate the exploration stagnation bottleneck in current VLA-RL frameworks and reveal that trajectory diversity is fundamentally more important to sample efficiency than the sheer quantity of collected rollouts. 
    Motivated by these insights, we introduce RL Exploration Token (ExToken), a simple yet general framework that condition VLA policies on discrete behavioral priors derived from offline demonstrations for structured exploration. By conditioning the policy on different tokens during rollout collection, ExToken encourages the agent to explore diverse behavioral modes, substantially improving state-action coverage and exploration efficiency. To bridge exploration during training with deterministic inference at deployment, ExToken further incorporates a state-conditioned token selector that adaptively predicts effective behavioral modes for unseen scenarios. Extensive experiments across simulated and real-world robotic manipulation tasks demonstrate that ExToken consistently accelerates convergence, improves task performance, and exhibits strong robustness under highly constrained interaction budgets.

%% file: Section/1_Intro.tex
\vspace{-\baselineskip}
\begingroup
    \makeatletter
    \renewcommand{\tableofcontents}{\@starttoc{toc}}
    \makeatother

    \hypersetup{linkcolor=BerkeleyBlue}

    \titlecontents{section}[0em]
        {\vspace{4pt}\large\bfseries}
        {\thecontentslabel\quad}       
        {}                             
        {\hfill\contentspage}          

    \setcounter{tocdepth}{1}
    \tableofcontents
\endgroup 

\newpage
\section{Introduction}

Vision-Language-Action (VLA)~\citep{kim2024openvla,black2024pi_0,intelligence2025pi_} models have recently demonstrated strong capabilities in robotic manipulation, and reinforcement learning (RL) has further emerged as a promising paradigm for post-training VLAs through environmental interaction and human feedback~\citep{lu2025vla,liu2026can,li2025simplevla,chen2025pirl,intelligence2025pi,chen2025conrft} (see Appendix~\ref{related} for a broader review).
However, the practical deployment of RL for VLAs is hindered by prohibitive sample inefficiency. Subjecting large-scale VLA architectures to extensive trial-and-error in physical or simulated environments incurs excessive interaction costs and safety risks, establishing sample efficiency as a primary bottleneck for scaling VLA post-training.

A growing body of VLA-RL algorithms have been developed which can implicitly improve sample efficiency. Most of these approaches adopt offline-to-online learning paradigms~\citep{xu2024rldg,guo2025improving,ghasemipour2025self} or focus on refining reward formulation and value functions~\citep{zhai2025vision,fei2025srpo,chen2025tgrpo} to more fully leverage the available data. Although such strategies yield notable benefits, they inherently overlook the fundamental relationship between active exploration and sample efficiency.

To this end, we first revisit the conventional exploration mechanisms employed during VLA-RL fine-tuning. We observe that exclusively depending on standard stochastic sampling gradually drives the policy toward action mode collapse~\citep{jin2025soe}, where the agent repeatedly generates highly redundant trajectories, leading to severe exploration stagnation. 
To understand the antidote to this issue, we investigate the intrinsic value of enforcing action diversity during rollouts. 
By simply discarding the most homogeneous half of the explored trajectories and retaining the most diverse half for training, we uncover a compelling insight: \textit{the diversity of the trajectories utilized for learning is fundamentally more critical to algorithmic performance and sample efficiency than the sheer quantity of the rollouts explored}. This naturally raises a critical question: how can we actively enhance the diversity of the explored trajectories?

Building upon these insights, we propose ExToken, a unified framework for structured exploration in VLA-RL fine-tuning. Instead of relying solely on unstructured stochastic exploration, ExToken conditions the policy on discrete behavioral priors derived from offline human demonstrations. Specifically, we employ a pre-trained video encoder~\citep{jian2025rzenembed}
to extract trajectory representations and perform clustering in the latent space, where each cluster centroid defines an exploration token corresponding to a distinct behavioral mode present in the demonstration data. During RL training, conditioning the policy on different tokens encourages the agent to generate diverse trajectories, substantially improving rollout diversity and state-action coverage.
To bridge structured exploration during training with deterministic inference at deployment, ExToken further incorporates a state-conditioned token selector. Given the initial observation, the selector adaptively predicts the most suitable token and is jointly optimized alongside the VLA policy throughout RL training. This design enables context-aware exploration while preserving stable inference behavior. Extensive experiments across both simulated and real-world robotic manipulation tasks demonstrate that ExToken consistently improves task performance, accelerates policy convergence, and exhibits strong robustness under highly constrained interaction budgets.

In summary, our research makes three significant contributions in the field of VLA-RL:  
\begin{enumerate}
    \item We rethink the challenge of RL in VLA from the perspective of exploration, analyze the action mode collapse with the RL process, and empirically demonstrate that trajectory diversity is fundamentally more important to sample efficiency than the sheer quantity of collected rollouts (Section~\ref{sec: investigation}).

    \item Based on these findings, we introduce RL Exploration Token (ExToken), a simple yet general framework that explicitly guides VLA policies to explore structured and diverse behavioral modes through discrete token conditioning, while incorporating a state-conditioned token selector for both context-aware exploration and deterministic inference. 
    (Section~\ref{sec:method}).

    \item Extensive experiments across simulated and real-world robotic manipulation tasks demonstrate that ExToken consistently improves exploration diversity, accelerates policy convergence, and achieves strong robustness and generalization under highly constrained interaction budgets (Section~\ref{sec:exp}).
    
\end{enumerate}

%% file: Section/2_Rethink.tex
\section{Investigating the Interplay between Exploration and Sample Efficiency}\label{sec: investigation}

\begin{figure}[t] 
    \raggedright
    \includegraphics[width=0.95\columnwidth]{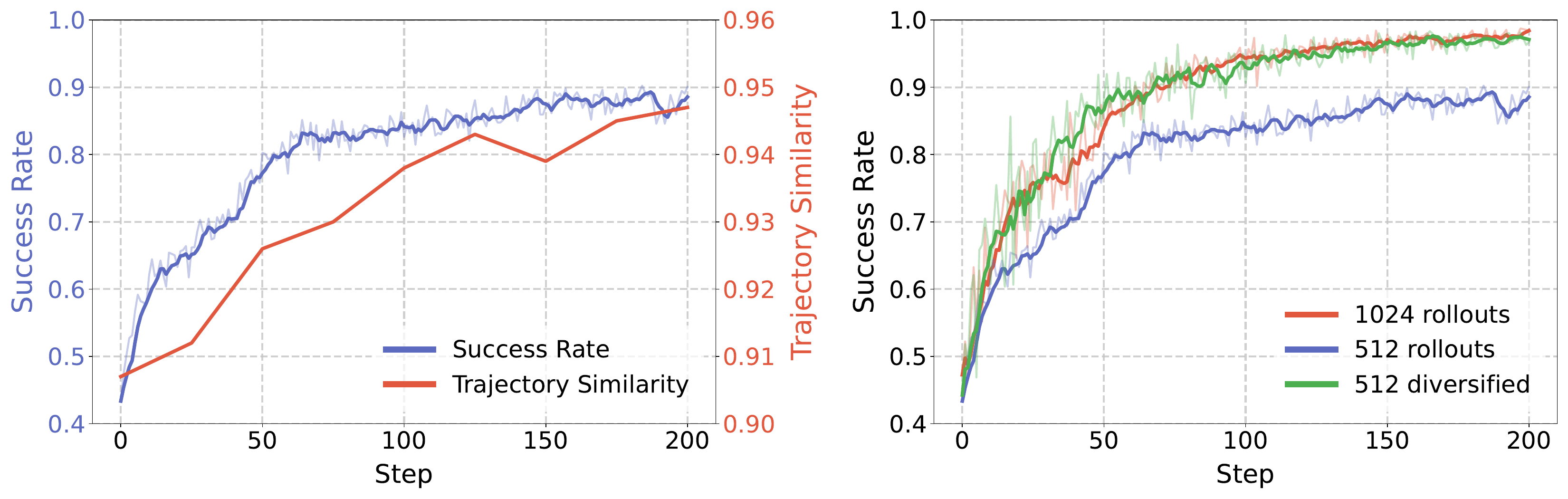} 
    \vspace{-5pt}
    \caption{(Left) As training progresses, trajectory similarity increases linearly, leading to premature convergence to a suboptimal performance level. (Right) Training with 512 diversified trajectories achieves comparable results to using 1024 standard rollouts.}
    \label{fig:motivation}
\end{figure}

\subsection{Exploration Stagnation during VLA-RFT}
To better understand the interplay between exploration dynamics and training efficiency in VLA-RL, we first investigate how exploratory behavior evolves throughout online RL fine-tuning. Standard VLA-RL paradigms typically rely on stochastic sampling to explore the environment. However, our empirical analysis reveals that as policy optimization progresses, the generated trajectories become increasingly homogeneous, causing the policy to repeatedly exploit similar action patterns, which is termed as action mode collapse.

Specifically, following prior VLA-RL settings, we adopt OpenVLA-OFT~\citep{kim2025fine} initialized from a single demonstration trajectory and further fine-tune it using RLinf-GRPO~\citep{zang2025rlinf} on LIBERO-Object~\citep{liu2023libero} (see Appendix~\ref{preliminary} for preliminaries). To quantify exploration diversity, we employ a pre-trained video encoder~\citep{jian2025rzenembed} to extract trajectory representations and compute pairwise  similarity. As illustrated in Figure~\ref{fig:motivation}(Left), trajectory similarity steadily increases throughout RL optimization, while task performance prematurely plateaus at a sub-optimal level. This observation suggests that unstructured stochastic exploration is insufficient to sustain effective exploration in the high-dimensional action spaces of VLA models.


\subsection{Trajectory Diversity Matters}
The emergence of action mode collapse naturally raises a pivotal question: Can enforcing behavioral diversity during exploration improve learning efficiency?
To answer this, we systematically analyze how algorithmic performance scales with respect to both the sheer volume of environmental interactions and the diversity of the explored trajectories. Specifically, we conduct an empirical study comparing three distinct configurations: (1) 1024 rollouts: Training with a large batch of standard exploratory trajectories at each RL step; (2) 512 rollouts: Training with a reduced batch size; (3) 512 diversified rollouts: Collecting 1024 rollouts during exploration, but discarding the most homogeneous half and only retaining the 512 most diverse trajectories for policy training.

Our experimental results in Figure~\ref{fig:motivation}(Right) yield a compelling insight: the policy trained on 512 diversified rollouts consistently matches the 1024-rollout baseline, while drastically outperforming the standard 512-rollout setting. 
These results reveal a key insight: \textit{trajectory diversity is fundamentally more important to sample efficiency than the sheer quantity of collected interactions.}
This finding shifts the focus of VLA-RL from merely collecting more data toward generating more diverse trajectories, motivating our central question: how can we explicitly encourage structured and diverse exploration during RL, without relying on expensive, brute-force oversampling?


%% file: Section/3_Method.tex
\section{ExToken: Structured Exploration for VLA-RL}\label{sec:method}

\begin{figure}[t] 
    \center
    \includegraphics[width=0.95\columnwidth]{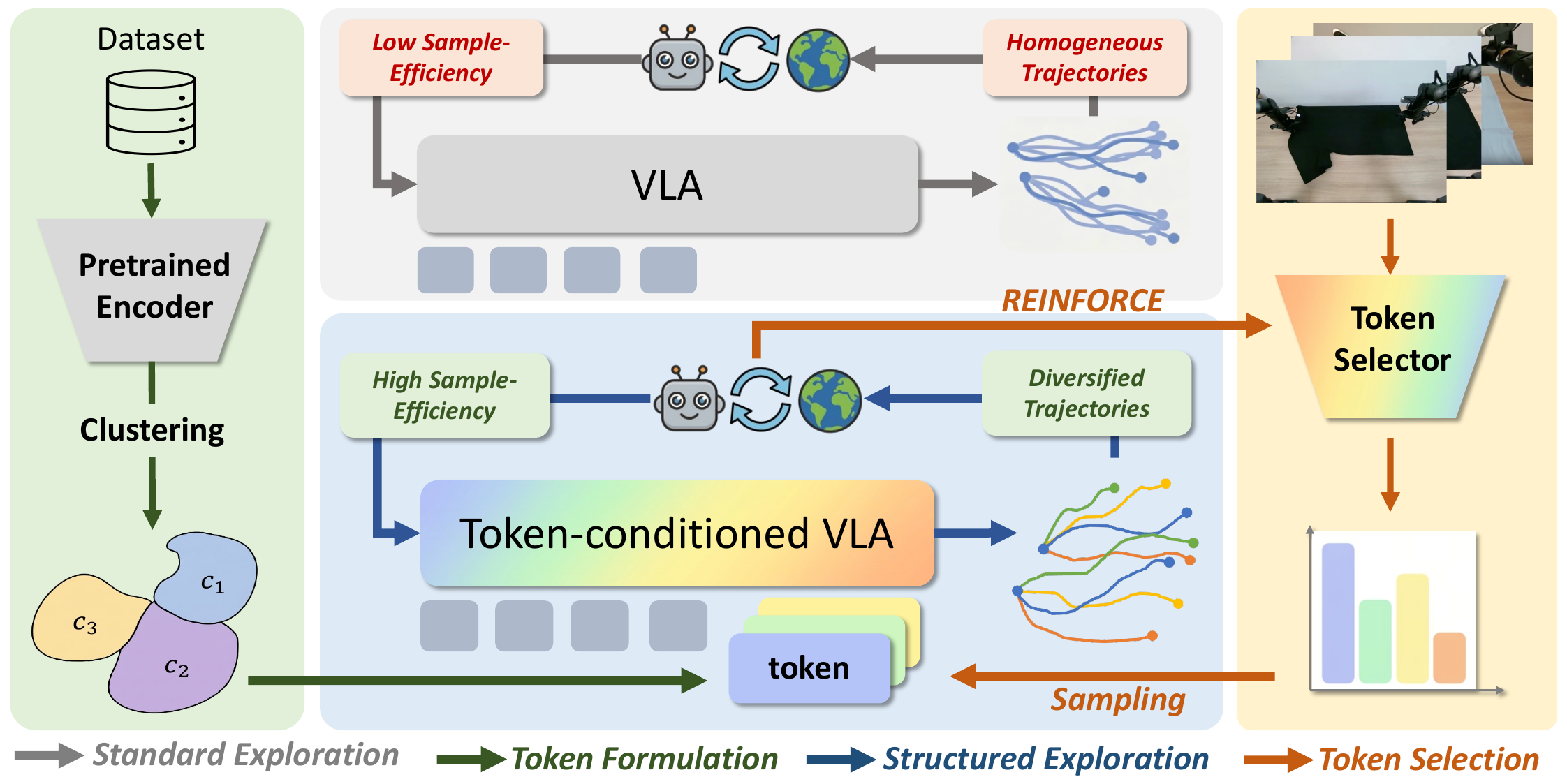
    } 
    \caption{Overview of ExToken. Compared to standard exploration, ExToken explicitly utilizes a token to steer structured exploration towards diverse trajectories, achieving higher sample efficiency.} 
    \label{fig:overview}
\end{figure}

Motivated by the preceding analysis, we introduce ExToken, a framework for structured exploration in VLA-RL. ExToken consists of two key components: (i) a token conditioning mechanism that guides the policy to explore diverse behavioral modes derived from offline demonstrations; and (ii) a state-conditioned Token Selector, which adaptively predicts the most suitable token for a given initial scenario.

\subsection{Promoting Structured Diversity via Exploration Token}\label{sec:token}
We first elaborate on the formulation of the specific exploration tokens, and subsequently demonstrate how they promote structured exploration during RL fine-tuning.

\textbf{Token Formulation.} 
Inspired by the rich diversity of behavioral modes inherently contained within expert demonstration datasets~\citep{yang2025discover}, we ground our exploration tokens directly in these demonstrations. Specifically, given a dataset $\mathcal{D}$, we employ a pre-trained video embedding model $\mathcal{E}$~\citep{jian2025rzenembed} to extract latent spatial-temporal features and perform $K$-means clustering in the latent space:
\begin{equation}
    \{c_1, c_2, \dots, c_K\} = K\text{-means}(\mathcal{E}(\mathcal{D})),
\end{equation}
where each centroid $c_k$ of the clustered embeddings  defines a discrete exploration token $k$ corresponding to a distinct behavioral mode, and $K$ denotes the total number of such tokens..

\textbf{Token-Conditioned Warm-Up.} 
To enable the VLA policy to interpret and condition its behavior on these tokens, we directly append the token to the end of the input sequence within the token embedding layer, and conduct a supervised fine-tuning (SFT) warm-up phase. Specifically, each demonstration trajectory is paired with a specific token, which is the center corresponding to its assigned cluster, allowing the policy to learn the association between different tokens and their corresponding behavioral modes. This initialization phase aligns different tokens with their corresponding trajectory behaviors, allowing the policy to generate token-conditioned action patterns during subsequent RL exploration.

\begin{wrapfigure}[13]{r}{0.4\textwidth}
    \vspace{-15pt} 
    \centering
    \includegraphics[width=\linewidth]{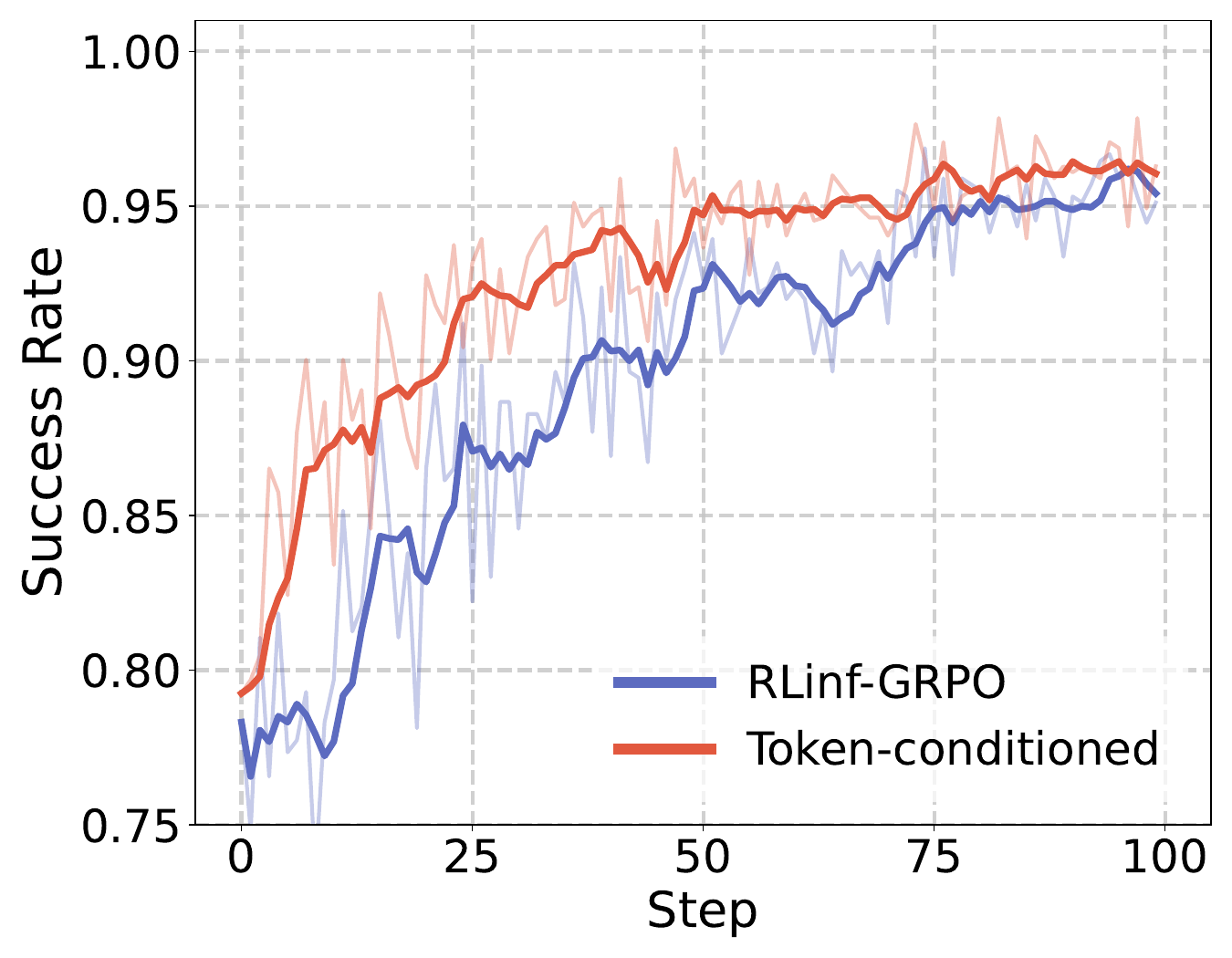}
    \vspace{-20pt}
    \caption{Token-conditioned structured exploration improves sample efficiency compared to baseline on LIBERO-Long.}
    \label{fig:method_token}
    \vspace{-8pt} 
\end{wrapfigure}
\textbf{Structured Exploration during RL.} 
During the RL interaction phase, we promote exploration diversity by conditioning each rollout on a uniformly sampled exploration token. Given an initial state $s_0$, we sample $\text{token}_k \sim \mathcal{U}(1, K)$ and prepend it to the policy input. Unlike conventional stochastic exploration mechanisms, these tokens explicitly guide the policy toward different behavioral modes, substantially improving rollout diversity and state-action coverage during training.
As illustrated in Figure~\ref{fig:method_token}, introducing token-conditioned exploration significantly reduces trajectory homogenization throughout RL optimization, thereby facilitating accelerated convergence and superior final task performance under limited interaction budgets. However, while randomized token conditioning is effective for exploration during training, it cannot be directly applied during deployment, since the optimal token for an unseen scenario is unknown a priori.




\subsection{Reinforce Token Selection through ExToken}\label{sec:framework}
To bridge the gap between training exploration and deterministic inference, we introduce ExToken, a framework that jointly optimizes a state-conditioned Token Selector alongside the VLA policy.

\textbf{State-Conditioned Token Selection.}
Given the initial observation $s_0$ consisting of an image observation $o_0$ and a language instruction $l$, the Token Selector $\phi$ predicts a categorical distribution over the token space. Specifically, we employ a pre-trained SigLIP encoder~\citep{zhai2023sigmoid} followed by an MLP head to predict the logits of each token:
\begin{equation}
P_\phi(\text{token}_k \mid s_0; \tau)=
\frac{\exp\!\big(z_k / \tau\big)}{\sum_{j=1}^{K}\exp\!\big(z_j / \tau\big)},
\qquad
z = \text{MLP}\big(\text{SigLIP}(s_0)\big)\in\mathbb{R}^{K},
\end{equation}
where $\tau$ denotes the sampling temperature. We warm up the Token Selector on the clustered offline dataset using the assigned cluster labels as supervision, enabling it to acquire an initial state-to-token mapping. During RL, the tokens are sampled from the temperature-scaled distribution to maintain exploration diversity while progressively biasing selection toward state-relevant behavioral modes. During inference, the selector deterministically outputs the most likely token via
$
\arg\max_{\text{token}_k} P_\phi(\text{token}\mid s_0)
$.


\textbf{REINFORCE Bi-level Optimization.} 
During RL training, the Token Selector $\phi$ and VLA policy $\pi_\theta$ are optimized jointly. While the VLA policy is updated using standard RL objectives, the Token Selector is optimized at the episode level as a single-step decision process using REINFORCE~\citep{williams1992simple}:
\begin{equation}
\nabla_\phi J(\phi)=
\sum_{k=1}^{K}
P_\phi(\text{token}_k \mid s_0)
\left(R(\tau)-b\right)
\nabla\phi
\log
P_\phi(\text{token}_k \mid s_0),
\end{equation}
where $R(\tau)$ denotes the trajectory reward and $b$ is a baseline for variance reduction. Through this joint optimization process, the selector progressively learns to predict effective behavioral modes for different initial states as policy learning evolves.



Notably, ExToken is designed as a lightweight and broadly compatible exploration framework. Since our proposed tokens are introduced solely through input conditioning, the framework can be seamlessly integrated into different VLA architecture on both online RL and iterative offline RL settings to promote structurally diverse exploration.

%% file: Section/4_Exp.tex
\section{Experiments and Analysis}\label{sec:exp}
In this section, we conduct extensive experiments to answer the questions: (1) Does ExToken improve sample efficiency under limited interaction budgets? (2) Does ExToken promote more diverse exploration behaviors? (3) Is ExToken robust to interaction constraints and token granularity?

\subsection{Simulation Experiment}
\textbf{Settings.}
To first validate our method on online RL, we conduct experiments on four LIBERO task suites (Spatial, Object, Goal, and Long), each comprising 10 distinct tasks. We adopt OpenVLA-OFT as the base policy and optimize it using GRPO within the RLinf framework.  We set the number of exploration tokens to $K=6$ across all experiments.  We first construct exploration tokens from the offline demonstrations and perform token-conditioned SFT warm-up, using 6 trajectories per cluster for LIBERO-Long and 2 trajectories per cluster for the remaining suites, followed by online RL to iteratively collect rollouts and update the policy within the simulator. To evaluate sample efficiency under constrained interaction budgets, we limit rollout collection to 512 trajectories per optimization step and train for 100 RL steps. More implementation details are provided in the Appendix~\ref{experiment}. 

\textbf{Baselines.}
We compare ExToken against both supervised and RL-based VLA baselines, including GRAPE~\citep{zhang2024grape} (DPO~\citep{rafailov2023direct} on OpenVLA), VLA-RL~\citep{lu2025vla} (PPO~\citep{schulman2017proximal} on OpenVLA), TGRPO~\citep{chen2025tgrpo} (GRPO~\citep{shao2024deepseekmath} on OpenVLA), SimpleVLA-RL~\citep{li2025simplevla} and RLinf-GRPO~\citep{zang2025rlinf} (GRPO on OpenVLA-OFT). To demonstrate the sample efficiency gains, RLinf-GRPO uses the same warm-up data configuration and RL interaction budget as ExToken.


\begin{table}[t]
\centering
\caption{Main results on LIBERO benchmark~\citep{liu2023libero}, where $^\dagger$ indicates results produced under our constrained interaction budget setting.}
\label{tab:main_results_simulation}
\small
\begin{tabularx}{\columnwidth}{l c >{\centering\arraybackslash}X >{\centering\arraybackslash}X >{\centering\arraybackslash}X >{\centering\arraybackslash}X >{\centering\arraybackslash}X}
\toprule
\textbf{Models} & \textbf{Paradigm} & \textbf{Spatial} & \textbf{Object} & \textbf{Goal} & \textbf{Long} & \textbf{Average} \\
\midrule

$\pi_0$~\citep{black2024pi_0}  & SFT & 96.8 & 98.8 & 95.8 & 85.2 & 94.2 \\
$\pi_{0.5}$~\citep{intelligence2025pi_}  & SFT & 98.8 & 98.2 & 98.0 & 92.4 & 96.9 \\
OpenVLA~\citep{kim2024openvla}  & SFT & 84.7 & 88.4 & 79.2 & 53.7 & 76.5 \\
OpenVLA-OFT~\citep{kim2025fine}  & SFT & 97.6 & 98.4 & 97.9 & 94.5 & 97.1 \\
\midrule
GRAPE~\citep{zhang2024grape}  & RL & 88.5 & 92.1 & 83.1 & 57.2 & 80.2 \\
TGRPO~\citep{chen2025tgrpo}  & RL & 90.4 & 92.2 & 81.0 & 59.2 & 80.7 \\
VLA-RL~\citep{lu2025vla}  & RL & 90.2 & 91.8 & 82.2 & 59.8 & 81.0 \\
SimpleVLA-RL~\citep{li2025simplevla}  & RL & 98.2 & 98.7 & \textbf{98.8} & 91.7 & 96.9 \\
RLinf-GRPO$^\dagger$~\citep{zang2025rlinf}  & RL & 98.6 & 98.4 & 95.1 & 95.2 & 96.8 \\
\rowcolor{gray!15} ExToken (Ours)$^\dagger$ & RL & \textbf{99.0} & \textbf{99.4} & 96.5 & \textbf{97.8} & \textbf{98.2} \\
\bottomrule
\end{tabularx}
\end{table}
\begin{figure}[t] 
    \centering
    \includegraphics[width=\columnwidth]{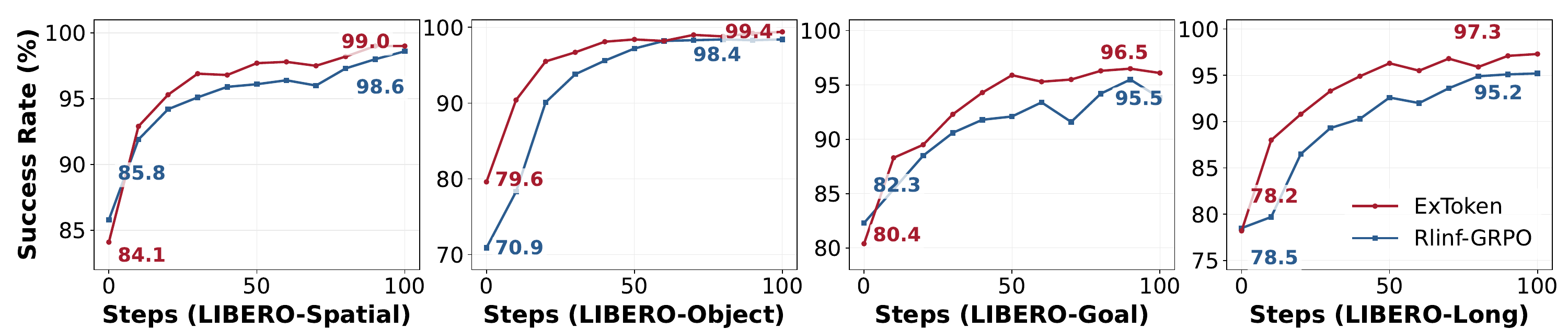} 
    \caption{Learning curves of \textcolor[HTML]{A61C2E}{\textbf{ExToken}} and \textcolor[HTML]{2B5C8F}{\textbf{RLinf-GRPO}} across the four LIBERO task suites, illustrating success rates (\%) over online RL fine-tuning steps.}
    \label{fig:learning_curves}
\end{figure}

\textbf{Results.}
Table~\ref{tab:main_results_simulation} demonstrates the quantitative comparison across the LIBERO benchmark. ExToken achieves the superior performance among the OpenVLA- and OpenVLA-OFT-based RL baselines. Under the exact same warm-up protocol and restricted RL interaction budget, ExToken achieves an overall average success rate of 98.2\%, outperforming the same setting  baseline RLinf-GRPO (96.8\%). 
The improvement is particularly evident on the challenging LIBERO-Long suite, where ExToken improves the success rate from 95.2\% to 97.8\%, suggesting that structured exploration leads to more effective policy optimization under constrained interaction budgets.
ExToken also outperforms strong SFT baselines inlucing $\pi_{0.5}$ (96.9\%) and OpenVLA-OFT (97.1\%), which heavily rely on the substantial expert demonstration dataset. Furthermore, Figure~\ref{fig:learning_curves}  shows that ExToken consistently converges faster than RLinf-GRPO across all four task suites, indicating that introducing exploration tokens to explicitly encourage structured action diversity helps alleviate learning stagnation and improve sample efficiency throughout RL training.

\subsection{Real-World Experiment}

\textbf{Settings.} We further evaluate ExToken on four complex real-world robotic manipulation tasks: \textit{Fold clothes}, \textit{Wipe table with towel}, \textit{Pour water}, and \textit{Insert pen into pen holder}. The latter three tasks are co-trained under a multi-task learning setting. Following recent iterative offline RL methods~\citep{intelligence2025pi,evorl2026} for real-world VLAs, we adopt $\pi_{0.5}$ as the base model and build upon the Evo-RL framework~\citep{evorl2026}. 
During iterative RL training, we intentionally constrain the rollout collection budget to evaluate learning efficiency in limited-interaction settings. Specifically, we perform two RL iterations for  \textit{Fold clothes} and a single iteration for the multi-task suite, collecting 20 rollouts per task in each iteration. To facilitate the online data collection, we employ human-in-the-loop (HIL) intervention~\citep{luo2025precise}, allowing for seamless corrective feedback during the limited exploration.

\begin{table*}[t]
\centering
\newcommand{\graygap}[1]{\textsubscript{\textcolor{gray!80}{\tiny(#1)}}} 

\caption{\textbf{Comparison on Real-World Tasks.} Evaluation includes the original training scenario and three generalization settings: unseen objects, background variations (-BG), and lighting conditions. All methods are evaluated over 20 rollouts under varied positions.}
\label{tab:real_world_comparison}

\setlength{\tabcolsep}{3.5pt} 
\small
\begin{tabularx}{\textwidth}{l | *{4}{>{\centering\arraybackslash}X} | *{4}{>{\centering\arraybackslash}X}}
\toprule
\multirow{2}{*}{\textbf{Methods}} & \multicolumn{4}{c|}{\textbf{Fold clothes}} & \multicolumn{4}{c}{\textbf{Wipe table with towel}} \\
\cmidrule(lr){2-5} \cmidrule(lr){6-9}
& Original & \mbox{-Object} & \mbox{-BG} & \mbox{-Lighting} & Original & \mbox{-Object} & \mbox{-BG} & \mbox{-Lighting} \\
\midrule
$\pi_{0.5}$~\citep{intelligence2025pi_} & 75 & 65\graygap{-10\%} & 70\graygap{-5\%} & 55\graygap{-20\%} & 70 & 55\graygap{-15\%} & 60\graygap{-10\%} & 60\graygap{-10\%} \\
Evo-RL~\citep{evorl2026}      & 90 & 70\graygap{-20\%} & 80\graygap{-10\%} & 65\graygap{-25\%} & 90 & 80\graygap{-10\%} & 85\graygap{-5\%} & 85\graygap{-5\%} \\
ExToken        & 95 & 90\graygap{-5\%}  & 90\graygap{-5\%}  & 80\graygap{-15\%} & 95 & 85\graygap{-10\%} & 90\graygap{-5\%} & 90\graygap{-5\%} \\
\midrule
\midrule
\multirow{2}{*}{\textbf{Methods}} & \multicolumn{4}{c|}{\textbf{Pour water}} & \multicolumn{4}{c}{\textbf{Insert pen into pen holder}} \\
\cmidrule(lr){2-5} \cmidrule(lr){6-9}
& Original & \mbox{-Object} & \mbox{-BG} & \mbox{-Lighting} & Original & \mbox{-Object} & \mbox{-BG} & \mbox{-Lighting} \\
\midrule
$\pi_{0.5}$~\citep{intelligence2025pi_} & 65 & 50\graygap{-15\%} & 55\graygap{-10\%} & 45\graygap{-20\%} & 65 & 45\graygap{-20\%} & 50\graygap{-15\%} & 45\graygap{-20\%} \\
Evo-RL~\citep{evorl2026}      & 80 & 70\graygap{-10\%} & 65\graygap{-15\%} & 70\graygap{-10\%} & 85 & 65\graygap{-20\%} & 75\graygap{-10\%} & 70\graygap{-15\%} \\
ExToken        & 90 & 85\graygap{-5\%}  & 80\graygap{-10\%} & 75\graygap{-15\%} & 90 & 75\graygap{-15\%} & 85\graygap{-5\%} & 80\graygap{-10\%} \\
\bottomrule
\end{tabularx}
\end{table*}

\textbf{Results.} As illustrated in Table~\ref{tab:real_world_comparison}, ExToken consistently outperforms the competitive Evo-RL baseline across all tasks in "original" setting, achieving an average improvement of 6.25\%. Given the restricted interaction budget (only 20 rollouts per iteration), this confirms that ExTokens provides highly effective directional guidance, allowing the policy to converge to a higher performance ceiling than standard iterative RL. In addition, while the strong $\pi_{0.5}$ and Evo-RL baseline suffer severe performance degradation when encountering environmental variations, ExToken demonstrates resilience. 
As indicated by the bracketed decay metrics, the performance drop of ExToken is typically limited to $5\%-10\%$, whereas the baselines often experience substantially larger degradation under the same perturbations. We attribute this improvement to the structured diversity induced by token-conditioned exploration, which encourages the policy to experience more varied interaction patterns during RL training instead of repeatedly exploiting homogeneous behaviors. As a result, the learned policy generalizes more effectively to unseen real-world variations.


\subsection{Further Analysis}
\begin{figure}[t] 
    \raggedright
    \includegraphics[width=\columnwidth]{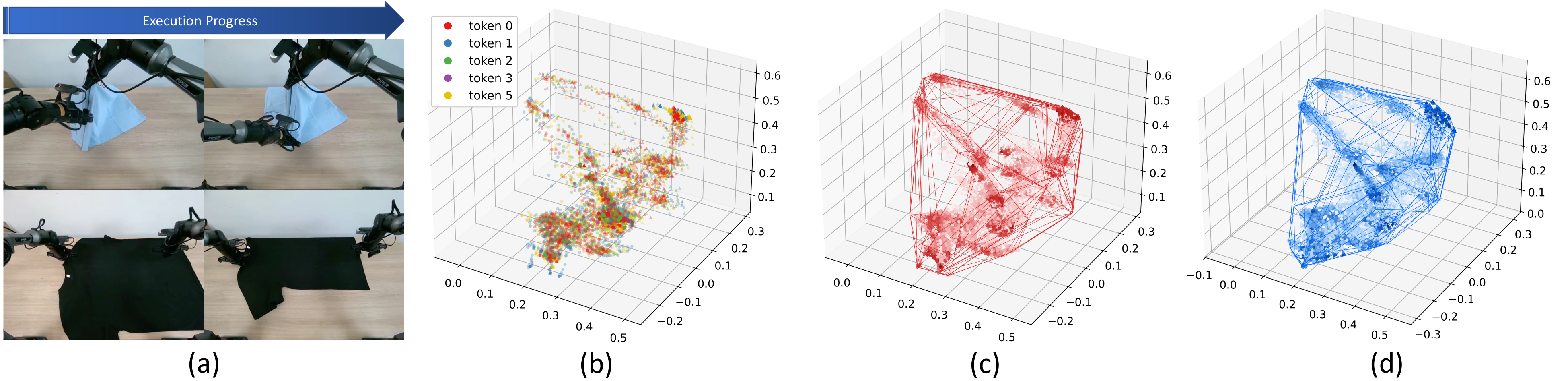} 
    \caption{Token clustering analysis and exploration space comparison.}
    \label{fig:analyse}
\end{figure}

\textbf{Do ExToken Clusters Capture Distinct Behaviors?}
To better understand the role of ExTokens, we analyze the behavioral properties of the generated token clusters. By clustering latent representations extracted from demonstration trajectories, ExTokens capture distinct behavioral modes within the offline dataset. Figure~\ref{fig:analyse}(a) visualizes two representative clusters from \textit{Fold clothes}. One cluster primarily corresponds to trajectories where the garment is initially misaligned, leading the policy to perform corrective straightening behaviors before folding. Another cluster captures trajectories that involve post-folding smoothing behaviors to refine the final cloth configuration. These examples suggest that the learned tokens encode consistent behavioral structure from the demonstrations.

Importantly, the effectiveness of ExToken does not depend on every token corresponding to a perfectly human-interpretable behavior. Due to the complexity and continuity of robotic manipulation, some clusters may reflect subtle variations that are difficult to semantically characterize. Instead, the primary role of ExTokens is to provide diverse conditioning signals during rollout collection, encouraging the policy to generate varied  behaviors across episodes. We find that such structured behavioral variation  is sufficient to provide the exploration diversity required during RL training.

\textbf{Can ExToken Promote Diverse Exploration Trajectories?}
Beyond the semantic rationale and structural discrimination of the clustered tokens, the fundamental premise of ExToken is its ability to translate these discrete tokens into diverse physical rollouts. To verify whether the VLA model genuinely conditions its behavior on the provided tokens and expands its exploration space, we visualize the 3D end-effector (EE) pisition generated during the RL exploration phase.

As illustrated in Figure~\ref{fig:analyse}(b), conditioning the policy on different sampled tokens produces clearly different trajectory patterns. The resulting EE positions form distinct spatial regions, indicating that the policy generates token-conditioned interaction behaviors rather than repeatedly sampling homogeneous rollouts. To further quantify exploration diversity, we construct convex hulls over the explored EE positions. Compared with standard stochastic exploration (Figure~\ref{fig:analyse}(d)), token-conditioned exploration (Figure~\ref{fig:analyse}(c)) covers a substantially larger spatial region, which suggests that the tokens encourage the policy to explore a broader range of behaviors during RL training, thereby improving exploration diversity and alleviating exploration stagnation.

\begin{wrapfigure}{r}{0.4\textwidth}
    \vspace{-20pt} 
    \centering
    \includegraphics[width=\linewidth]{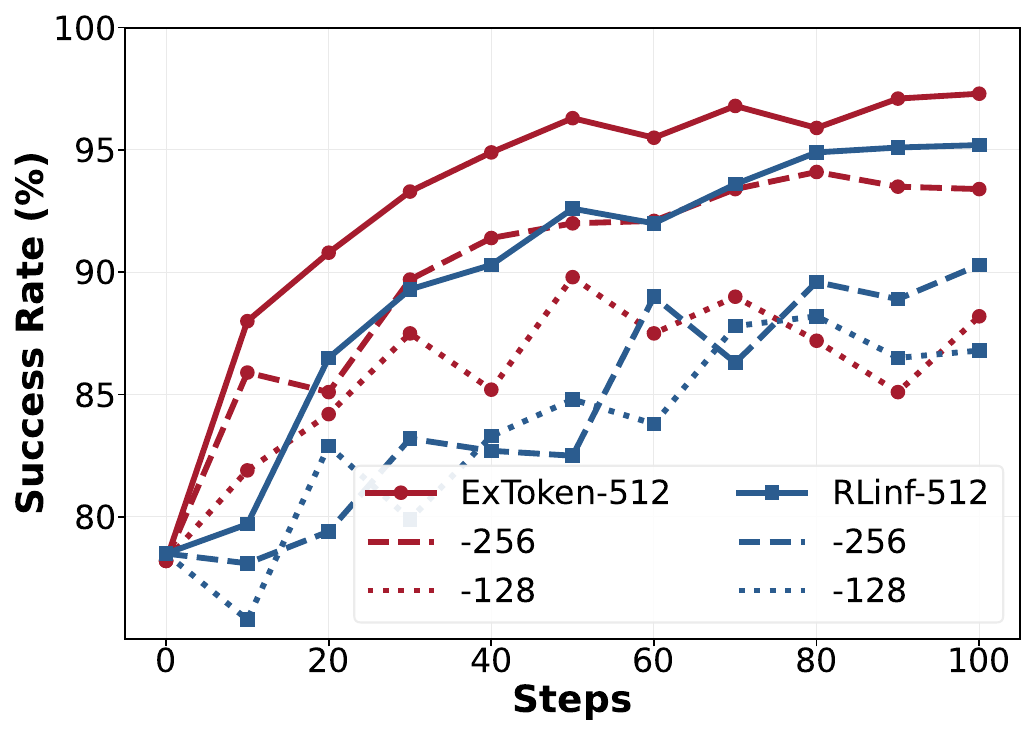}
    \vspace{-25pt}
    \caption{Comparison across varying numbers of exploration rollouts.}
    \label{fig:rollout}
    \vspace{-10pt} 
\end{wrapfigure}
\textbf{Can ExToken Alleviate Extreme Interaction Constraints?}
A paramount challenge in Embodied AI is minimizing the prohibitive cost of environmental interactions. To evaluate the resilience under extreme constraints, we reduce the rollout budget per optimization step from 512 to 128 trajectories. As shown in Figure~\ref{fig:rollout}, standard stochastic exploration is highly sensitive to reduced rollout budgets. RLinf-GRPO drops to 90.3\% success rate at 256 rollouts and exhibits unstable training dynamics at 128 rollouts. In contrast, ExToken demonstrates stronger robustness and graceful degradation, as our method utilizing only 256 rollouts (93.4\%) directly rivals the RLinf baseline using double the data budget (512 rollouts). However, ExToken also becomes less stable at the extreme limit of 128 rollouts. We attribute this behavior to increased gradient variance caused by sampling diverse tokens from a very limited number of initial states. Adaptively reducing the cluster count under such restricted budgets presents a promising avenue.


\begin{wrapfigure}{r}{0.4\textwidth}
    \vspace{-15pt} 
    \centering
    \includegraphics[width=\linewidth]{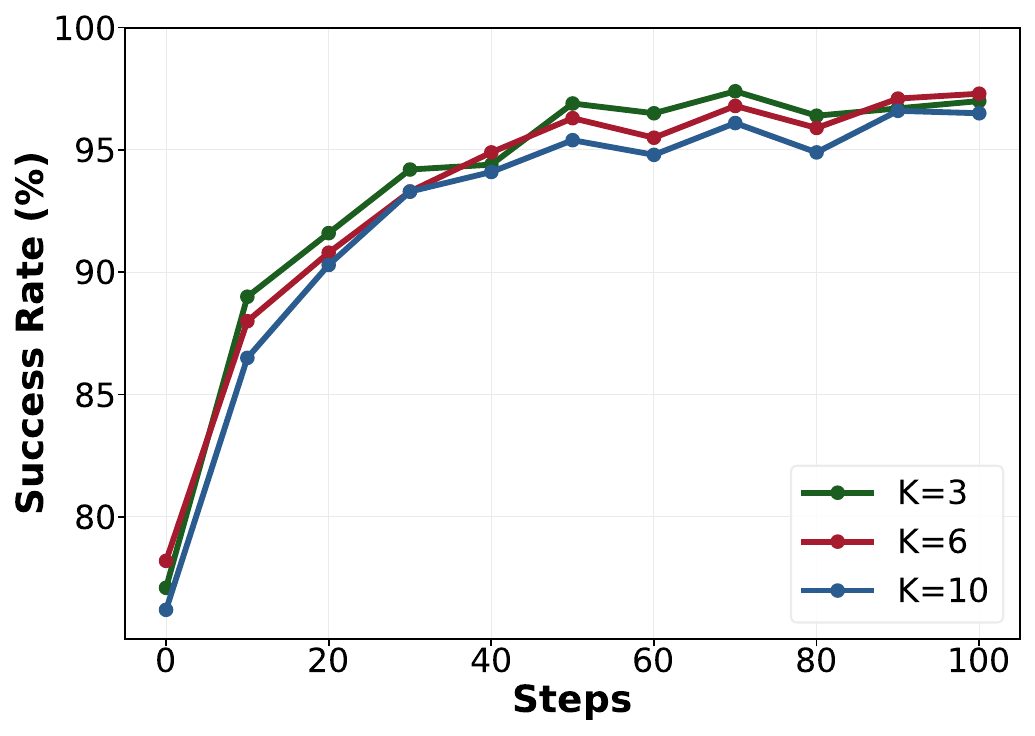}
    \vspace{-25pt}
    \caption{Comparison across different token granularity $K$.}
    \label{fig:ablation_k}
    \vspace{-10pt} 
\end{wrapfigure}
\textbf{Is ExToken Robust to Token Granularity?}
Finally, we study the sensitivity of ExToken to the number of token clusters $K$. We evaluate the framework with $K \in {3, 6, 10}$ and observe relatively stable performance across different settings. In particular, the results remain highly comparable between $K=3$ and $K=6$, while a slight performance drop appears at $K=10$. We hypothesize that excessively fine-grained token partitioning may increase the difficulty of learning stable token-conditioned behaviors. Overall, these results suggest that ExToken is generally robust to token granularity, ensuring reliable exploration without rigorous tuning $K$.



%% file: Section/5_Conclusion.tex
\section{Conclusion and Future Work}

In this work, we first identified exploration stagnation and action mode collapse as an important bottleneck hindering RL for VLA models, and reveal  maintaining trajectory diversity is fundamentally more critical to unlocking sample efficiency than merely scaling data volume. Based on these insights, we introduced the RL Exploration Token (ExToken), a simpel yet general framework for structured exploration that guides VLA toward diverse behavioral modes through token-conditioned interaction. By leveraging behavioral priors extracted from offline demonstrations, ExToken promotes diverse and structured exploration during RL while remaining compatible with different VLA architectures and RL paradigms. Extensive experiments in both simulated and real-world robotic tasks demonstrate that ExToken consistently improves policy performance, accelerates convergence, and enhances robustness under constrained interaction budgets and environmental variations.

While ExToken significantly enhances sample efficiency, our framework adopts a simplified design to validate the core hypothesis of token-guided exploration, presenting several avenues for future refinement. First, the current utilization of tokens is relatively basic. A natural extension is to jointly optimize the token representations  with the policy, or introduce probabilistic token dropout to unconditionally capture trajectory diversity and implicitly ensemble performance. Second, the current tokens are derived from clustering video representations extracted by a pre-trained encoder. Adopting more structured approaches, such as state-conditioned latent models, may yield exploration tokens that better capture task dynamics. Finally, ExToken conditions the policy using a single token predicted solely from the initial state. While effective for short-horizon tasks, extending this to fine-grained, temporally-conditioned tokens would allow the agent to dynamically shift exploration strategies for complex, long-horizon manipulation tasks.



%% file: Section/Appendix.tex


\input{Appendix/1_related}

\input{Appendix/2_preliminary}

\input{Appendix/3_experimental}

%% file: Appendix/1_related.tex
\section{Related Work}\label{related}

\subsection{Reinforcement Fine-tuning for VLA models}

Reinforcement learning (RL)~\citep{sutton1998reinforcement,qing2022survey} has recently been adopted as an effective post-training technique for vision-language-action (VLA) models, enabling policies to improve beyond imitation learning~\citep{qing2024a2po,qing2025bitrajdiff,su2023uncertainty} through online trial-and-error interactions~\citep{xia2024deer,kong2024tptu,wang2026language}.
Early works mainly focus on adapting existing RL algorithms~\citep{kong2025mastering,liu2025curricular,zhou2023centralized} to the VLA setting. 
For example, VLA-RL~\citep{lu2025vla} extend the classical online RL algorithm PPO~\citep{schulman2017proximal} to VLA online post-training, while RL4VLA~\citep{liu2026can} further introduce LLM-oriented RL algorithms, such as GRPO~\citep{shao2024deepseekmath} and DPO~\citep{rafailov2023direct}, into VLA fine-tuning to investigate the advantages of RL for VLAs. 
Meanwhile, systems such as SimpleVLA-RL~\citep{li2025simplevla} and RLinf-VLA~\citep{zang2025rlinf} provide open-source training frameworks that facilitate the deployment of RL fine-tuning for VLA models. These studies establish the feasibility of RL-based policy improvement for embodied tasks. Building on these paradigms, recent methods make further efforts to improve RL performance. 
$\pi_{RL}$~\citep{chen2025pirl} reformulates the iterative denoising process in flow-based VLA models to allow for exact action log-likelihood computation; 
TGRPO~\citep{chen2025tgrpo} extends GRPO from sparse trajectory-level advantage to dense step-wise advantage signals for fine-grained policy updates;
SRPO~\citep{fei2025srpo} uses the model's own successful rollouts as self-references to extract valuable learning signals from failed attempts; and model-based methods such as VLA-RFT~\citep{li2025vla} and WMPO~\citep{zhu2025wmpo} leverage world models to enable safer and efficient interaction with environments for policy improvement. In addition, latent-planning methods~\citep{chen2026last, huang2026thinkact} augment action prediction with future-state latents, encouraging policies to account for long-term reasoning. 
While these methods improve RL fine-tuning for VLA models, they  fundamentally focus on improving policy optimization, credit assignment, planning, or sample utilization given collected interactions, leaving the role of exploration itself  underexplored. Existing methods typically rely on the policy’s inherent stochasticity to generate new experiences. In constrast, our work addresses this complementary direction by explicitly enhancing trajectory diversity to improve RL exploration and sample efficiency.

\subsection{Exploration in Robot Learning}
Active exploration aims to design efficient exploration policies that guide agents to generate action samples with higher performance gains, thereby mitigating the sample inefficiency introduced by random rollouts in conventional robot learning. Most existing exploration approaches can be categorized into two main lines. (1) Reward-based methods~\citep{parisi2021interesting,ecoffet2019go,burda2018large,pathak2017curiosity} promote exploration by introducing intrinsic rewards—typically based on novelty or curiosity—to incentivize the agent to visit unfamiliar or unpredictable states. (2) Sampling-based methods diversify action generation through strategies such as epsilon-greedy~\citep{mnih2013playing,van2016deep}, Boltzmann exploration~\citep{szepesvari2022algorithms}, and goal-directed exploration~\citep{hu2023planning}, which are generally achieved by perturbing the action space or injecting noise into the policy. Beyond these,
SIME~\citep{jin2025sime} achieves safe and structured exploration by injecting stochastic perturbations into latent representations rather than directly perturbing generated actions.
SOE~\citep{jin2025soe} extends this idea by learning a task-relevant latent state and introducing stochastic perturbations within this latent space, enabling safe and efficient exploration.
However, the above methods are developed for relatively small-scale models.
RESample~\citep{xue2025resample} first learns a  coverage function and then actively explores the coverage boundary, collecting targeted “deviation-and-recovery” interaction trajectories to improve VLA performance.
PLD~\citep{xiao2025self} introduces a lightweight residual RL framework together with a hybrid execution mechanism to autonomously generate  corrective trajectories.
Although RESample and PLD extend exploration capabilities in the VLA setting, both primarily operate as data augmentation approaches and cannot be directly applied to online reinforcement learning.


%% file: Appendix/2_preliminary.tex
\section{Preliminaries}\label{preliminary}
We formulate the Vision-Language-Action (VLA) model as a parameterized policy $\pi_{\theta}$ that maps multimodel inputs $s_t$ (i.e. visual observations $o_t$ and language instructions $l$) into an action chunk $a_{t:t+H}$, where $t$ denotes the timestep and $H$ denotes the action horizon. The policy can be optimized through supervised fine-tuning (SFT) using offline expert demonstrations or via reinforcement learning (RL) via environment interactions.


\textbf{RL Formulation.} 
We formulate robot control as a Markov decision process (MDP) $(\mathcal{S}, \mathcal{A}, p, r, \mathcal{\gamma})$, where $\mathcal{S}$ is the observation space, $\mathcal{A}$ is the action space, $p(s_{t+1}|s_t, a_t)$ denotes the transition dynamics of the environment, $r(s_t, a_t)$ is the reward function, and $\gamma\in [0,1)$ is the discount factor. The goal of RL is to learn a policy $\pi_{\theta}$ that maximizes the expected return $\mathcal{J}_{RL}=\mathbb{E}_{\tau\sim\rho_{\pi}}[\sum_{t=0}^T \gamma^t r_t]$, where $\rho_{\pi}(\tau)$ denotes the trajectory distribution induces by the policy $\pi$. We assume access only to a sparse binary reward at the end of each episode as success or failure.

\textbf{Group Relative Policy Optimization (GRPO)} is an online RL method eliminates the value function by computing advantages through group-relative normalization. Given an initial state $s_0$, the behavior policy $\pi_{old}$ generates $G$ trajectories $\{\tau_i\}_{i=1}^G$. The objective can be formulated as:
\begin{equation}\small
    \mathcal{J}_{GRPO}(\theta) = \mathbb{E}_{s_0\sim\mathcal{D}, \{\tau_i\}\sim\pi_{old}} \left[ \frac{1}{G} \sum_{i=1}^G \frac{1}{|\tau^{(i)}|} \sum_{t=1}^{|\tau^{(i)}|} \min \left( \rho_t^{(i)}(\theta)\hat{A}^{(i)}, \text{clip}(\rho_t^{(i)}(\theta), 1-\epsilon, 1+\epsilon)\hat{A}^{(i)} \right) \right].
\end{equation}
The importance sampling ratio $\rho_t^{(i)}$ and the normalized advantage $\hat{A}^{(i)}$ are defined as:
\begin{equation}
    r_{i,t}(\theta) = \frac{\pi_\theta(a_{i,t}|s_{i,t})}{\pi_{\theta_{\text{old}}}(a_{i,t}|s_{i,t})}, \quad \hat{A}_i = \frac{R_i - \text{mean}(\{R_i\}_{i=1}^G)}{\text{std}(\{R_i\}_{i=1}^G)},
\end{equation}
where $R_i$ denotes the total reward of the $i$-th trajectory, $\epsilon>0$ denotes the PPO clipping parameter that limits the policy ratio.

\textbf{RL with Experience and Corrections via Advantage-conditioned Policies (RECAP)} is an offline RL framework for VLA models that extracts improved policies through advantage conditioning. Given a dataset $\mathcal{D}_{\pi_{\mathrm{ref}}}$ collected by a reference policy $\pi_{\mathrm{ref}}$, RECAP first trains a value function on collected trajectories to estimate state-action advantages, and then performs policy extraction through advantage-conditioned supervised learning.

Based on the advantage-conditioned policy improvement formulation, RECAP learns a unified policy model that supports both unconditional action generation $\pi_\theta(a_t|o_t,\ell)$, and advantage-conditioned action generation $\pi_\theta(a_t|I_t,o_t,\ell)$:

\begin{equation}
\mathcal{J}_{\mathrm{RECAP}}(\theta)=
\mathbb{E}_{(o_t,a_t,\ell)\sim\mathcal{D}{\pi_{\mathrm{ref}}}}
\left[
-\log \pi_\theta(a_t|o_t,\ell)
-\alpha \log \pi_\theta(a_t|I_t,o_t,\ell)
\right],
\end{equation}

where $\alpha$ is the trade-off between behavior cloning and advantage-conditioned policy extraction.

By conditioning on the binary advantage indicator, RECAP can leverage both demonstrations and autonomous rollouts, including suboptimal trajectories, while avoiding explicit policy-gradient optimization. The resulting policy approximates an improved policy that favors actions with positive advantages and can be iteratively refined using newly collected experience.

%% file: Appendix/3_experimental.tex
\section{Experimental Details}\label{experiment}
\subsection{Details on LIBERO}
Our policy adopts the OpenVLA-OFT~\citep{kim2025fine} model simplified by SimpleVLA-RL~\citep{li2025simplevla}, and is initialized with pre-trained OpenVLA~\citep{kim2024openvla} weights. To construct our warm-up dataset, we first extract trajectory-level embeddings from the open-source LIBERO dataset using the Rzen video encoder~\citep{jian2025rzenembed}, followed by K-means clustering. Given the multi-task setting of LIBERO, we perform clustering independently for each task in each suite to prevent the results from being biased by inter-task variations. We then randomly sample trajectories from each cluster: specifically, 6 trajectories per cluster for the LIBERO-Long suite, and 2 per cluster for the remaining three suites. All trajectories within the same cluster are assigned identical exploration tokens. Finally, we conduct Supervised Fine-Tuning (SFT) for 30K steps on 4 $\times$ A800 GPUs, adopting the same configurations as SimpleVLA-RL.

For the RL phase, our VLA training adopts the same configurations as the open-source RLinf-GRPO~\citep{zang2025rlinf}, utilizing 8 $\times$ A800 GPUs. We collect 512 rollouts per step and cap the total training steps at 100. The Token Selector is optimized via the REINFORCE algorithm every 5 steps, utilizing all rollouts accumulated during this interval for the update. The reward of a predicted token given an initial state is defined as the average reward of the trajectories within its group in GRPO. The baseline is similarly computed on a per-task basis. The REINFORCE learning rate is set to 1e-6, with a sampling temperature of 2.0.

\begin{figure}[t] 
    \center
    \includegraphics[width=\columnwidth]{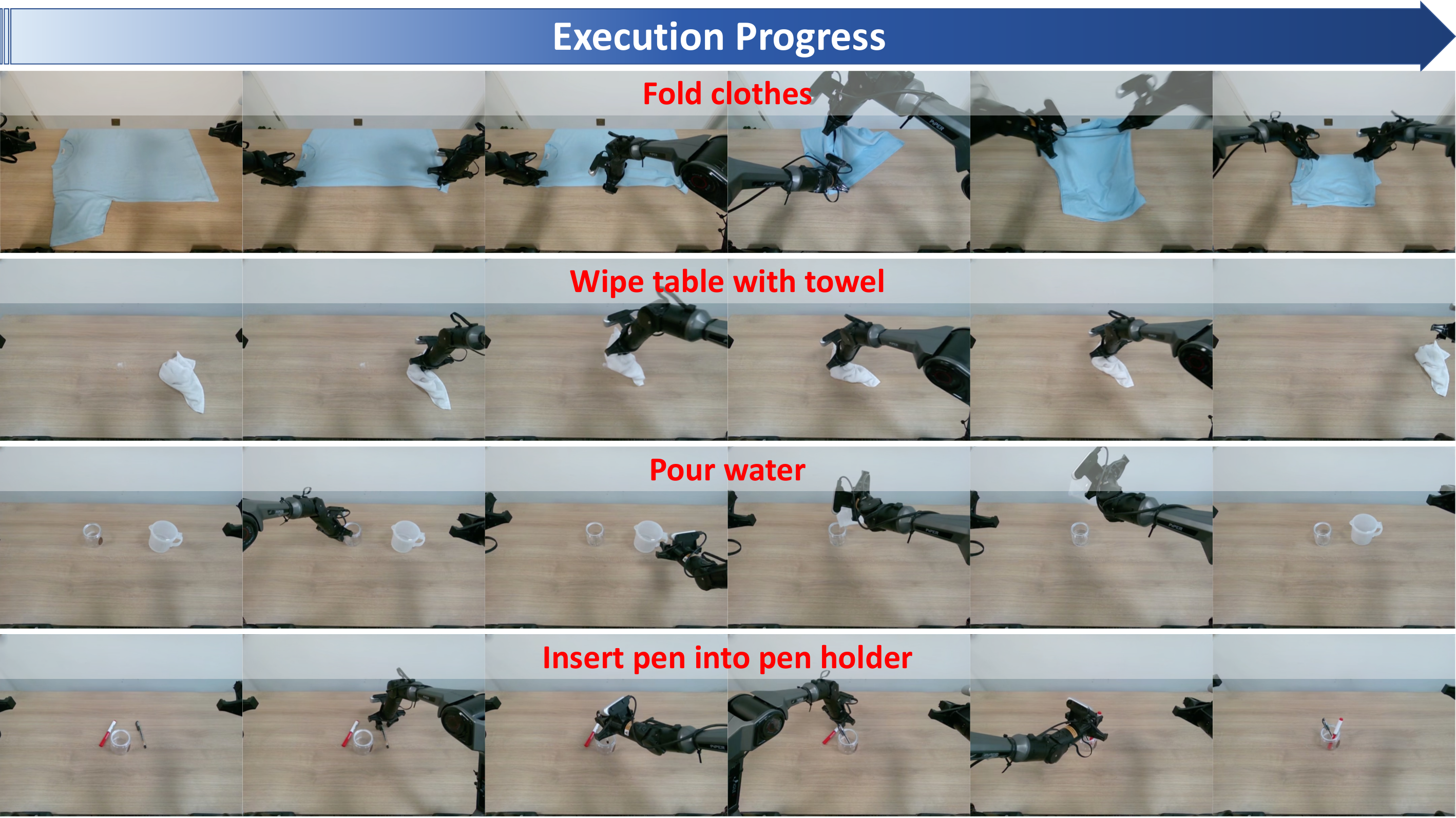}
    \vspace{-10pt}
    \caption{Overview of real-world execution trajectories. The sequence illustrate the continuous execution progress across diverse robotic manipulation tasks.} 
    \label{fig:realworld}
    \vspace{-15pt}
\end{figure}
\subsection{Details on Real-World}
\begin{figure}[t] 
    \center
    \includegraphics[width=\columnwidth]{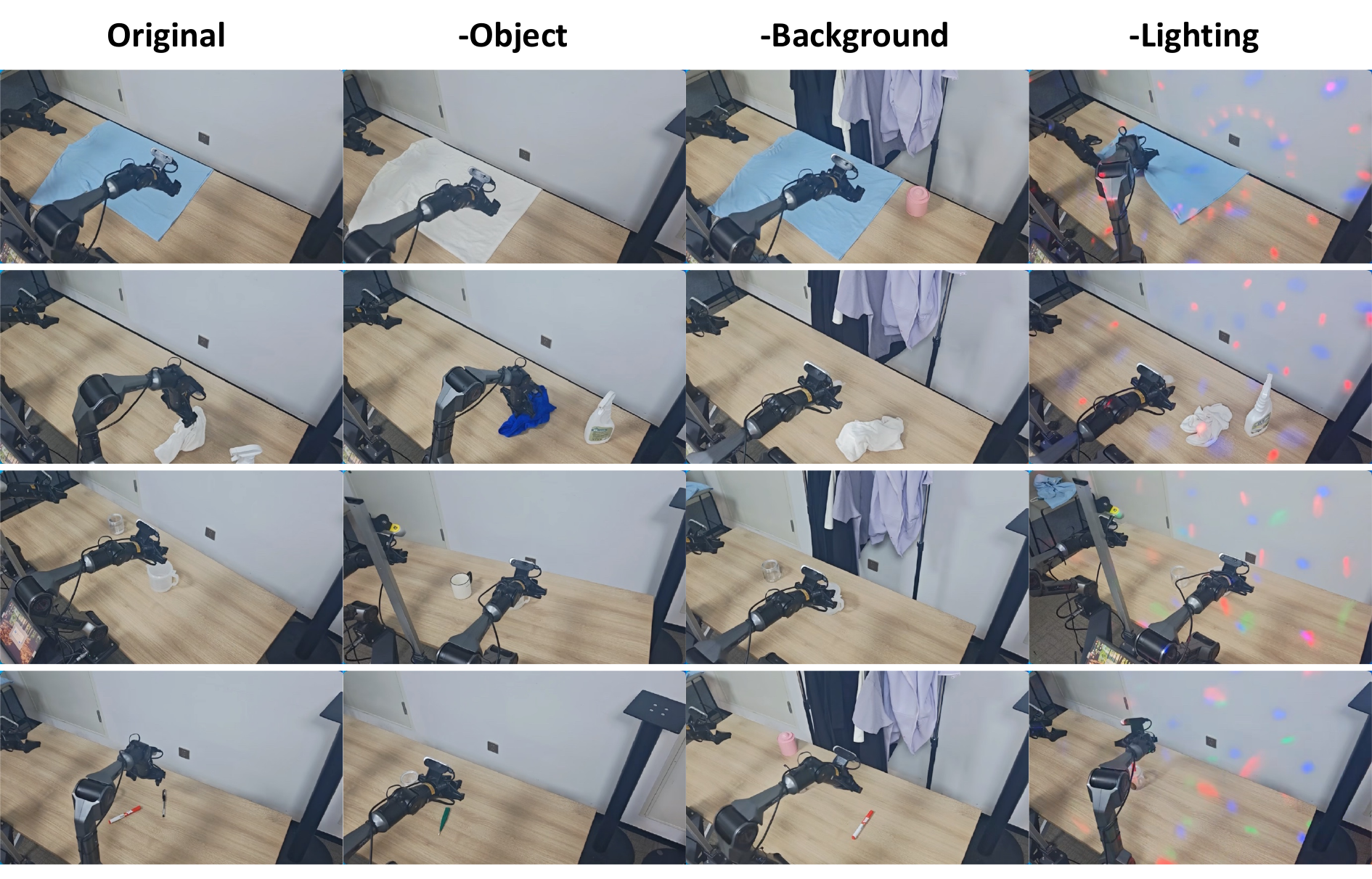}
    \vspace{-10pt}
    \caption{Policy robustness under unseen settings. The model consistently maintains success under severe visual variations.} 
    \label{fig:generalization}
\end{figure}

Our real-world experimental evaluations are conducted on the Cobot Magic platform, equipped with four AgileX Piper robotic arms: two leader
arms for human teleoperation and two follower arms for data collection and policy inference. We evaluate four real-robot tasks: (1) \textit{fold clothes}: folding a shirt and aligning it neatly; (2) \textit{wipe table with towel}: picking up the wiping cloth and cleaning stains on the table; (3) \textit{pour water}: inserting the gripper through the cup handle, lifting the cup, and pouring all water to a target container without spillage; (4) \textit{insert pen into pen holder}: picking up multiple pens and inserting them into a pen holder. All the objects are initialized at arbitrary locations. The visualization of the tasks and further generalization settings are illustrated in Figure~\ref{fig:realworld} and Figure~\ref{fig:generalization}.

We leverage Evo-RL~\citep{evorl2026} to implement value training and infer, as well as the Human-in-the-Loop iterative rollouts collection process, while model training uses the RECAP~\citep{intelligence2025pi} implemented on the official Openpi JAX codebase\footnote{\url{https://github.com/Physical-Intelligence/openpi}}. For video encoding, we first convert the collected data into grayscale to mitigate the interference of real-world factors such as variations in color and illumination. For the \textit{fold clothes} task, we utilize 200 trajectories for the warm-up phase. For the multi-task setting, we utilize 100 trajectories per task. The warm-up phase and all subsequent training iterations are conducted for 20K steps each. Throughout both stages, we utilize a peak learning rate of 2.5e-5 for the warm-up and 2.5e-7 for the subsequent iterations, applying a cosine decay schedule with a minimum learning rate ratio of 0.1. The token selector is trained every iteration in real-world tasks. All training is conducted on 8 $\times$ A800 GPUs, while for inference, the model is deployed on a single A800 GPU.

